\documentclass[10pt,twocolumn,letterpaper]{article}
\pdfoutput=1 
\usepackage{cvpr}              % To produce the CAMERA-READY version

\pdfoutput=1 

\usepackage{amsthm}
\usepackage{tabularx}
\usepackage{multirow}
\usepackage{algorithm}
\usepackage{algpseudocode}
\usepackage{amsmath}
\usepackage{xcolor}
\usepackage{soul} % For underlining
\usepackage[export]{adjustbox}  % For additional image 
\usepackage{trimclip}  % For proper croppingformatting options
\usepackage{longtable}
\usepackage[textsize=scriptsize]{todonotes} 

\pdfoutput=1 

\newcommand{\mymacro}[1]{#1}

\newcommand{\taxonomy}{\mymacro{\mathcal{T}}}

\newcommand{\node}{\mymacro{v}}

\newcommand{\hP}{\mymacro{\mathrm{\textbf{hP}}}}
\newcommand{\hR}{\mymacro{\mathrm{\textbf{hR}}}}
\newcommand{\hF}{\mymacro{\mathrm{\textbf{hF}}}}
\newcommand{\anc}{\mymacro{\mathrm{anc}}}

\newcommand{\nodes}{\mymacro{V}}
\newcommand{\labels}{\mymacro{\mathcal{L}}}

\newcommand{\rootnode}{\mymacro{\rho}}
\newcommand{\edges}{\mymacro{E}}

\newcommand{\gtnode}{\mymacro{\node_n^{\mathrm{gt}}}}
\newcommand{\prednode}{\mymacro{\node_n^{\mathrm{pr}}}}

\newcommand{\refnode}{\node}
\newcommand{\candnode}{\node^{\prime}}
\newcommand{\reference}{\mymacro{\labels(\refnode)}}
\newcommand{\candidate}{\mymacro{\labels({\candnode)}}}

\newcommand{\parent}{\mymacro{\mathrm{par}}}

\newcommand{\alphabet}{\mymacro{\Sigma}}
\newcommand{\stringset}{\mymacro{\alphabet^{*}}}

\definecolor{darkgreen}{rgb}{0.09, 0.45, 0.27}
\newcommand{\concept}[1]{{\color{darkgreen}\textsc{#1}}}

\definecolor{darkblue}{rgb}{0.1, 0.2, 0.6}
\newcommand{\modelpred}[1]{\textit{\color{darkblue}#1}}

\newcommand{\predvar}[1]{{\color{orange}\textit{#1}}}

\newcommand{\modelname}[1]{#1}

\definecolor{darkpurple}{rgb}{0.25, 0.15, 0.45}
\newcommand{\mtype}[1]{{\color{darkpurple}\textbf{#1}}}

\usepackage{amsmath}
\usepackage{bbm}
\usepackage{tabularx}
\usepackage{lineno}
\usepackage[accsupp]{axessibility}  % Improves PDF readability for those with disabilities.
\usepackage{xcolor}
\usepackage{soul}
\definecolor{mplblue}{RGB}{31, 119, 180}
\definecolor{mplyellow}{RGB}{255,127,14}
\usepackage[frozencache,cachedir=minted-cache]{minted}

\usepackage[super]{nth}
\setminted{linenos, firstnumber=last, escapeinside=@@, numbersep=4pt}

\usepackage{svg}
\definecolor{cvprblue}{rgb}{0.21,0.49,0.74}
\usepackage[pagebackref,breaklinks,colorlinks,allcolors=cvprblue]{hyperref}
\input{macros}

\def\paperID{16159} % *** Enter the Paper ID here
\def\confName{CVPR}
\def\confYear{2025}

\title{Taxonomy-Aware Evaluation of Vision--Language Models}
\author{Vésteinn Snæbjarnarson$^{1,2}$\thanks{Correspondence:  {\tt\href{mailto:vesteinnsnaebjarnarson@gmail.com}{vesteinnsnaebjarnarson@gmail.com}.}}
\qquad
Kevin Du$^{2}$
\qquad
Niklas Stoehr$^{2}$
\qquad
Serge Belongie$^{1}$ \\
Ryan Cotterell$^{2}$
\qquad
Nico Lang$^{1}$
\qquad
Stella Frank$^{1}$ \\[0.5em]
$^{1}$University of Copenhagen \quad $^{2}$ETH Zürich \\\\
}

\vspace{-5cm}

\begin{document}
\maketitle
\begin{abstract}
When a vision--language model (VLM) is prompted to identify an entity depicted in an image,
it may answer \modelpred{``I see a conifer,''} rather than the specific label \concept{norway spruce}.
This raises two issues for evaluation: Firstly, the unconstrained generated text needs to be mapped to the evaluation label space (i.e.,~\concept{conifer}).
Secondly, a useful classification measure should give partial credit to less-specific, but not incorrect, answers (\concept{norway spruce} being a type of \concept{conifer}).
To meet these requirements, we propose a framework for evaluating unconstrained text predictions such as those generated from a vision--language model against a taxonomy.
Specifically, we propose the use of hierarchical precision and recall measures to assess the level of correctness and specificity of predictions with regard to a taxonomy.
Experimentally, we first show that existing text similarity measures do not capture taxonomic similarity well.
We then develop and compare different methods to map textual VLM predictions onto a taxonomy.
This allows us to compute hierarchical similarity measures between the generated text and the ground truth labels.
Finally, we analyze modern VLMs on fine-grained visual classification tasks based on our proposed taxonomic evaluation scheme.
Data and code are made available at \url{https://github.com/vesteinn/vlm-eval}.\looseness=-1
\end{abstract}

\section{Introduction}
\label{sec:intro}
\begin{figure}[hptb]
    \centering
    \includesvg[width=0.9\linewidth]{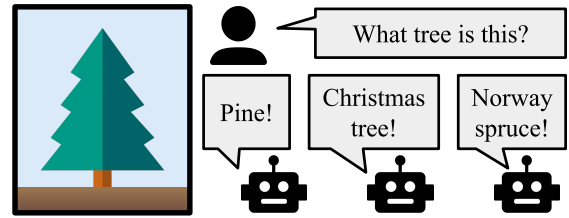}
    \caption{\textbf{Vision--language-models (VLMs) as fine-grained classifiers.} 
    VLMs generate text with varying degrees of specificity and similarity to gold-standard label classes. We tackle the problem of aligning these outputs to taxonomic classes.
    }
    \label{fig:hero}
    \vspace{-15pt}
\end{figure}
Modern vision–language models (VLMs)~\cite{openai_gpt-4_2023, liu2023llava, li2023blip2, fuyu-8b, chen2023minigptv2, NEURIPS2022_960a172b} serve as viable one-stop shops for a wide variety of tasks at the intersection of vision and language.
Given an image as input, VLMs generate unconstrained text, a key enabler of their flexibility across tasks. 
This flexibility, however, introduces additional challenges in their evaluation.
In this paper, we study the problem of evaluating the unconstrained text generated by VLMs in the context of fine-grained visual categorization (FGVC), and put forward two desiderata when evaluating VLMs on FGVC (\cref{fig:hero}):\looseness=-1
\begin{description}
\item[Evaluation with Textual Awareness.]
When tasked with classifying entities within an image, VLMs often generate text that does not inherently fit within the predefined set of
labels used in conventional evaluations. 
For example, a model might describe an image with the phrase \modelpred{pines in the snow}, while the ground-truth label is simply \concept{pine tree}. 
We suggest that a good evaluation should rely on fuzzy or representation-based matching techniques, which assess the similarity between the generated text and the predefined class labels, rather than whether the VLM generated the predefined class's label exactly. 
Such approaches have a long history in natural language processing (see, e.g.,~\cite{lin-2004-rouge,reimers_sentence-bert_2019}).
We call an evaluation measure for FGVC that is compatible with unconstrained text \defn{textually aware}.\looseness=-1

\begin{figure*}
    \centering
    \includegraphics[width=\linewidth]{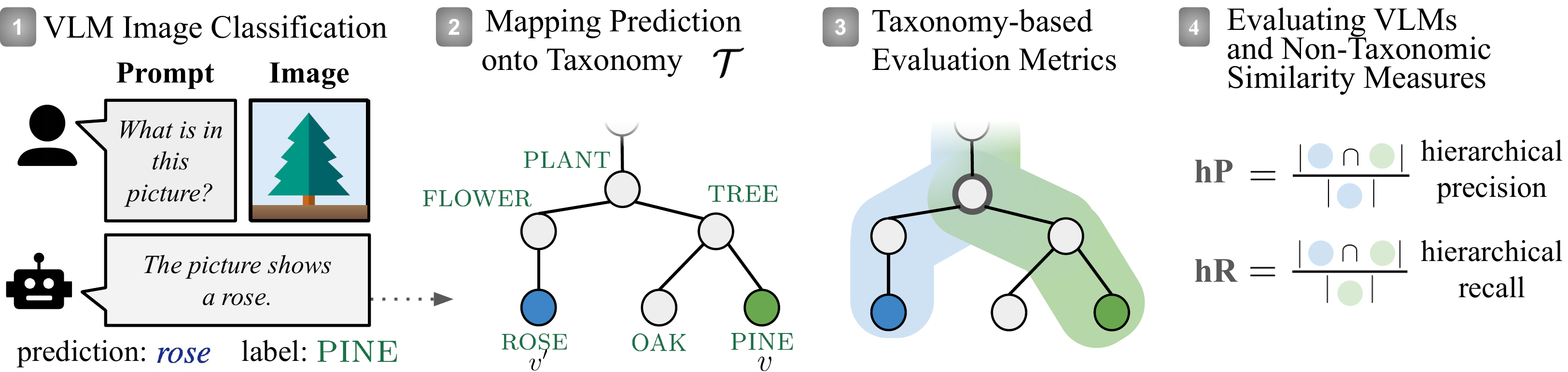}
    \caption{\textbf{Framework for taxonomy-aware VLM evaluation.} We propose a method for granular evaluation of the open-ended image classification capabilities of VLMs. To this end, we map model predictions onto a taxonomy, e.g., sourced from a large knowledge graph. We then use hierarchical precision ($\hP$) and recall ($\hR$) to evaluate how taxonomically accurate and specific a model prediction is. This, in turn, lets us evaluate and compare different VLMs and prompts.}
    \label{fig:overview}
\end{figure*}

\item[Evaluation with Hierarchical Awareness.]
Standard evaluation methods for VLM-based entity classification typically treat text generated by a VLM in a binary manner, i.e.,~it is either completely correct or incorrect \cite{deng2009imagenet, fgvcsurvey, hu2023open}.
However, many classification tasks involve label sets that are organized hierarchically, such as taxonomic trees or knowledge graphs. 
When such a structure is available, we suggest that evaluation measures take advantage of the structure, i.e., predictions should be judged based on their semantic proximity to the correct label within the hierarchy. 
Fortunately, these structured label spaces are common in FGVC and other real-world classification settings.
We call a VLM evaluation measure that is sensitive to taxonomic relationships \defn{hierarchically aware}.
\end{description}

In the technical portion of the paper, we propose a novel evaluation measure for assessing VLMs' performance on FGVC tasks that incorporates both textual and hierarchical awareness.
We term such an evaluation strategy \defn{taxonomy-aware}.
\cref{fig:overview} illustrates the approach for taxonomy-aware VLM evaluation that we present in this paper.
Our measure is based on hierarchical precision and recall \citep{Kiritchenko2006LearningAE} after placing strings generated by the VLM on a given taxonomy.
To contextualize our taxonomy-aware evaluation scheme within the literature, we first evaluate existing text similarity measures from NLP
to assess whether they implicitly capture taxonomic relationships.
Our analysis reveals that these measures do not align well with the underlying taxonomic structure, motivating our novel taxonomy-aware evaluation framework that explicitly maps unconstrained text generated from a VLM onto a taxonomy.\looseness=-1 

In the empirical part of the paper, we explore our taxonomy-aware evaluation scheme on a range of VLMs, varying in size, training strategies, and performance, under our proposed taxonomy-aware evaluation scheme.
To do so, we apply diverse prompt templates to control for accuracy, i.e., whether the model generates correct answers, and specificity, i.e., how detailed the response should be and whether to be more general when the model is not confident.
In support of our evaluation, we leverage large, carefully constructed taxonomies derived from knowledge graphs extracted from Wikidata~\cite{wikidata} and the Linnean Catalogue of Life~\cite{van2021benchmarking}.
These taxonomies span diverse domains, e.g.,~food, sports, animals, plants, cars, and landmarks, providing a holistic characterization of VLM behavior.
Our results demonstrate that our proposed taxonomy-aware evaluation measures effectively quantify varying degrees of specificity that more rigid evaluation measures like accuracy fail to capture.
We find that larger commercial models respond more successfully to prompts requesting varying levels of specificity,
revealing substantial differences compared to the prompt sensitivity exhibited by smaller, publicly available models.
These findings illustrate how taxonomy-aware evaluation can uncover nuanced model behaviors when conventional evaluation measures cannot.\looseness=-1

\section{Background}
We briefly review prior work on evaluating VLMs and textual similarity measures, with emphasis on approaches to visual categorization and hierarchical evaluation. A more general overview of the use of hierarchical labels for visual categorization is given in \cref{sec:sem_sim_hiearchy} in the supplement.

\subsection{Evaluating VLMs for Classification}
Recent work has explored VLM evaluation in open-domain visual recognition and question answering, often as a byproduct of general model development.
A dominant approach involves retrieving the most similar target class label to an unconstrained text using maximum inner-product search~\cite{conti2023vocabularyfree, hu2023open}. 
The retrieved label can then be directly used for classification or further processed to extract candidate labels. In some cases \cite{ging2024openended}, these extracted labels are compared to the original image using image-to-text similarity methods, such as CLIP~\cite{clip}.
In-context learning for classification~\cite{hakimov-schlangen-2023-images} leverages large language models and a few example labels to generate or extract an image label from a longer caption. Other approaches rely on models like GPT-4~\cite{openai_gpt-4_2023} to score generated answers based on a given question and reference answer~\cite{bai2023touchstone, zhu2023judgelm, kocmi-federmann-2023-large, liu2023llava}.
A recent method~\cite{ging2024openended} expands classification label spaces by prompting models to produce increasingly specific labels using taxonomically structured templates extracted from ImageNet~\cite{deng2009imagenet}. However, final model predictions are still evaluated using string-matching measures.
Some multi-task VLM benchmarks include image classification tasks~\cite{xu2023lvlmehub, bitton2023visitbench}, typically using exact-match evaluation or text-based measures borrowed from image captioning; see \cref{sec:sem_sim_hiearchy} in supplementary material. The Open-domain Visual Entity (OVEN) benchmark~\cite{hu2023open} relies on representation-based methods while other work avoids direct classification tasks altogether~\cite{pmlr-v162-zhou22n}.\looseness=-1

\subsection{Annotation and Prediction Similarity}
\label{sec:sim_methods}

Language generation tasks such as image captioning have traditionally been evaluated using measure based on string similarity, e.g., \mtype{ROUGE}~\cite{lin-2004-rouge}, \mtype{BLEU}~\cite{bleu}, \mtype{METEOR}~\cite{banerjee-lavie-2005-meteor} and \mtype{CIDER}~\cite{Vedantam_2015_CVPR}.
They primarily rely on surface form (string) overlap in prediction and ground truth labels, and as such, do not capture semantic similarities.
This led to a family of trained measures that rely on a language model that, given a pair of inputs, predicts a similarity score, typically based on the semantic match. Examples of these are \mtype{BERTScore}~\cite{bert-score}, and \mtype{SentenceBERT}~\cite{reimers_sentence-bert_2019}.
Causal and bi-directional language models have been shown to encode varied hierarchical semantic knowledge in human-interpretable metric spaces~\cite{petroni_language_2019,pan_large_2023,he_language_2024}. 
Natural language inference (\mtype{NLI}), also known as textual entailment~\cite{bowman_large_2015, wang_entailment_2021}, can also capture hierarchical specificity judgments, since sub-categories entail their supercategories, i.e.,~\modelpred{This is a field sparrow} entails \modelpred{This is a bird}.
In computer vision, this idea has been adopted as visual entailment~\cite{xie2019visual}.

Contrastive vision and language models~\cite{clip, siglip} can be used to compare image and text pairs using top-$k$ retrieval by inner product similarity.
This method relies on the intrinsic quality of the representations. While impressive, these have been shown to fail in cases requiring combining complex compositional information that was not seen together at training time \cite{yuksekgonul2023when}. 
More recently, prompt-based methods for scoring textual similarity have been proposed:
For instance, the LLaVA models \cite{liu2023llava} are evaluated on the COCO\cite{10.1007/978-3-319-10602-1_48} image captioning datasets by asking GPT-4 \cite{openai_gpt-4_2023} to provide numerical values (1-10) for a range of measures such as relevance and accuracy. 
Another approach is fine-tuning large language models to act as evaluation judges~\cite{zhu2023judgelm}.
This approach has also been claimed to outperform other evaluation methods for machine translation~\cite{kocmi-federmann-2023-large}. A more detailed description of the text similarity measures we compare against in this paper is given in the supplement in section \cref{app:text_sim_measures}.\looseness=-1

\section{Taxonomy-aware Evaluation}
\label{sec:formalism}

\definecolor{labelblue}{RGB}{61, 133, 198}
\definecolor{prediction}{RGB}{106, 168, 79}
\definecolor{metricsgray}{RGB}{108, 122, 137}
\begin{figure*}[t]
\centering
    \begin{subfigure}{0.17\textwidth}       \includegraphics[width=\textwidth,height=0.85\textwidth,keepaspectratio=false,clip]{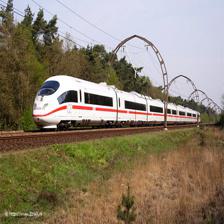}
        \caption*{ {\normalsize \concept{Train}}\par\medskip
        {\color{prediction}\textbf{Prediction:}}\par
        \hspace{0.5em}\modelpred{A mode of transport}\par\medskip
        {$\hP$: \textbf{1.00}} \hspace{3pt}
        {$\hR$: \textbf{0.75}}}
    \end{subfigure}
    \begin{subfigure}{0.17\textwidth}
\includegraphics[width=\textwidth,height=0.85\textwidth,keepaspectratio=false,clip]{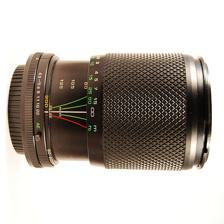}
        \caption*{ {\normalsize \concept{Telephoto lens}}\par\medskip
        {\color{prediction}\textbf{Prediction:}}\par
        \hspace{0.5em}\modelpred{A lens cover}\par\medskip
        {$\hP$: \textbf{0.80}} \hspace{3pt} 
        {$\hR$: \textbf{0.67}}}
    \end{subfigure}
    \begin{subfigure}{0.17\textwidth}
    \includegraphics[width=\textwidth,height=0.85\textwidth,keepaspectratio=false,clip]{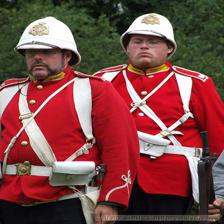}
        \caption*{ {\normalsize \concept{Pith helmet}}\par\medskip
        {\color{prediction}\textbf{Prediction:}}\par
        \hspace{0.5em}\modelpred{Neckwear}\par\medskip
        {$\hP$: \textbf{0.75}} \hspace{3pt} 
        {$\hR$: \textbf{0.50}}}
    \end{subfigure}
 \begin{subfigure}{0.17\textwidth}
        \includegraphics[width=\textwidth,height=0.85\textwidth,keepaspectratio=false,clip]{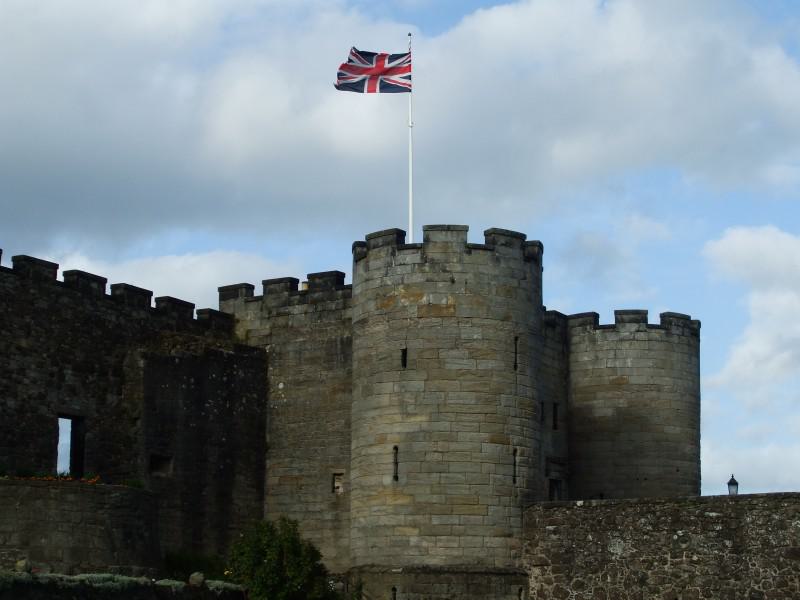}
        \caption*{ {\normalsize \concept{Stirling Castle}}\par\medskip
        {\color{prediction}\textbf{Prediction:}}\par
        \hspace{0.5em}\modelpred{Château de Ranrouët}\par\medskip
        {$\hP$: \textbf{0.83}} \hspace{3pt} 
        {$\hR$: \textbf{0.83}}}
    \end{subfigure}
     \begin{subfigure}{0.17\textwidth}
        \includegraphics[width=\textwidth,height=0.85\textwidth,keepaspectratio=false,clip]{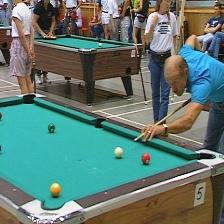}
        \caption*{ {\normalsize \concept{Pool}}\par\medskip
        {\color{prediction}\textbf{Prediction:}}\par
        \hspace{0.5em}\modelpred{High jump}\par\medskip
        {$\hP$: \textbf{0.67}} \hspace{3pt}
        {$\hR$: \textbf{0.67}}}
    \end{subfigure}
    \caption{\textbf{Illustrative examples.} Hierarchical precision ($\hP$) and recall ($\hR$) calculations on random label pairs (\cref{sec:evalsimmeasures}). While $\hP$ penalizes incorrect labels, i.e.,~labels that are further away from the target's ancestor set, $\hR$ penalizes mistakes made higher up on the taxonomy.}
    \label{fig:predictions}
    \vspace{-15pt}
\end{figure*}

Taxonomy-aware classification evaluation is a subtle and understudied problem \cite{Kiritchenko2006LearningAE,kosmopoulos2015,riehl2023}.
It is necessary to assign partial credit to answers that share some taxonomic similarity, e.g., to answers that are within the correct sub-tree, even if they do not match the correct node exactly.
\defn{Hierarchical precision} ($\hP$) and \defn{hierarchical recall} ($\hR$)~\cite{Kiritchenko2006LearningAE} are two existing measures that suitably assign partial credit.
Both take the length of the longest common path from the root node of the taxonomy---equivalent to the size of the \defn{ancestor set}---and normalize by either the length of the predicted path ($\hP$) or the target path ($\hR$).

We now introduce $\hP$ and $\hR$ formally.
Let $\alphabet$ be an alphabet of symbols, and $\stringset$ be the Kleene closure, i.e.,~the set of all strings whose symbols are drawn from $\alphabet$.
We assume the VLM under consideration constitutes a distribution over $\stringset$ when it is conditioned on an image.
We then define a \defn{taxonomy} $\taxonomy$ as a triplet $(\nodes, \labels, \edges)$, where $\nodes$ represents a finite set of \defn{nodes}, $\labels \colon \nodes \rightarrow \stringset$ is a function that maps each node to a string, called a \defn{label}, and $\edges \subset \nodes \times \nodes$ is a set of \defn{directed edges} between pairs of nodes.
Additionally, we require that the directed graph $(\nodes, \edges)$ forms a directed tree with a unique root $\rootnode \in \nodes$.
The set of nodes encountered along the unique path from the taxonomy tree's root node $\rootnode$ to a given node $\node$ constitutes that node's ancestor set, which we denote $\anc(\node)$. 
Formally, for any node $\node \in \nodes$, let $\parent(\node)$ denote $\node$'s unique parent node in the tree; $\parent(\rootnode)$, i.e., with the root note $\rootnode$ as an argument, is undefined.
We then recursively define $\anc(\node)$ as:\looseness=-1
\begin{subequations}
\begin{align}
\anc(\node) &\defeq \{\rootnode\} \justification{base case; $\node = \rootnode$}\\
\anc(\node) &\defeq \{\node\} \cup \anc(\parent(\node)). \justification{inductive case; $\node \neq \rootnode$}
\end{align}
\end{subequations}

We now discuss how to compare ground-truth and predicted nodes on a taxonomy. 
Let $\{(\gtnode, \prednode)\}_{n=1}^{N}$ be a set of pairs of nodes where $\gtnode$ is the ground-truth node for the $n^{\text{th}}$ image and $\prednode$ is the node mapped
from the string generated from the VLM conditioned on the $n^{\text{th}}$ image.
(We describe a procedure to create this string--node mapping in \cref{sec:mapping}.)
We then define $\hP$ and $\hR$ over $\{(\gtnode, \prednode)\}_{n=1}^{N}$ as follows\looseness=-1
\begin{subequations}
\begin{align}
    \hP &\defeq \frac{1}{N} \sum_{n=1}^{N} \frac{ | \anc(\prednode) \cap \anc(\gtnode)|}{|\anc(\prednode)|}, \\
    \hR &\defeq \frac{1}{N} \sum_{n=1}^{N}\frac{| \anc(\prednode) \cap \anc(\gtnode)|}{|\anc(\gtnode)|}.
\end{align}
\end{subequations}
The hierarchical F1 ($\hF$) is then defined as the harmonic mean of $\hP$ and $\hR$,
\begin{equation}
    \hF \defeq \frac{2 \cdot \hP \cdot \hR}{(\hP + \hR)}.
\end{equation}
Both $\hP$ and $\hR$ fall in the interval $[0, 1]$.
The maximum is achieved when the target and predicted paths coincide. 
Less-than-perfect scores capture the extent to which the path deviates from the gold answer.\looseness=-1
\begin{description}
  \item [$\hP$] \defn{Hierarchical precision} captures the amount of incorrect information in the prediction,~i.e., deviations from the correct path, relative to the number of nodes shared with the correct path.\looseness=-1
  \item [$\hR$] \defn{Hierarchical recall} measures the amount of correct information in the prediction, i.e., how much the path from the root to the predicted node overlaps with the path from the root to the ground truth node. 
  Hierarchical recall thus penalizes missing coverage of the target path, especially early deviations. 
  This can be seen as measuring the specificity of the prediction, in particular when the $\hP$ is 1.
\end{description}
We give an example of the computation of $\hP$ and $\hR$ in \cref{fig:predictions}, sampled from the OVEN datasets. 
In the case of the picture of \concept{Train}, we can see how only predicting \concept{mode of transport} leads to lower $\hR$ since it lacks information, while $\hP$ remains 1 since there is no incorrect information. For the sport \concept{pool} and \concept{Stirling Castle}, we notice how both scores drop, as incorrect nodes are predicted---these responses contain both incorrect information and some partially correct information. In the case of \concept{Pith helmet}, we see that the prediction is incorrect yet not very specific, resulting in a lower $\hR$ than $\hP$. We also note that it is not possible to have a lower $\hP$ than $\hR$ if the ground truth nodes are leaf nodes.

\section{Taxonomically Linked Datasets}
\label{sec:datasets}
To investigate taxonomy-aware evaluation measures for text generated from a VLM, we consider two taxonomies linked to FGVC tasks. 
First, the Catalogue of Life taxonomy as contained in the iNaturalist21 dataset~\citep{van2021benchmarking} is a clean, expert-curated taxonomy that maps evolutionary relationships between species and contains exactly 10,000 leaf nodes. The second is a more general taxonomy constructed from the Wikidata~\cite{wikidata} knowledge graph, which links entities from FGVC datasets in the Open-domain Visual Entity (OVEN) benchmark~\cite{hu2023open}.\looseness=-1

\subsection{Extracting Taxonomies}
While the iNaturalist21 dataset includes a built-in taxonomy, this is not the case for other FGVC datasets. For those included in the OVEN collection, we utilize Wikidata---a large-scale knowledge graph derived from Wikipedia~\cite{wikidata}. Unlike taxonomies, knowledge graphs are not constrained to tree or even directed acyclic graph structures, which necessitates the extraction of a taxonomy that fits the formal definition given in \cref{sec:formalism}.
To construct such a tree from Wikidata, we rely on the \texttt{subclass of} relation. When multiple paths exist from the root to a given node in the taxonomy, we retain the longest path.
In the case of ties, we sample one of the tied paths uniformly at random. 
To ensure a rooted directed tree, we exclude higher-level abstract classes that introduce cycles. 
Because OVEN already links its datasets (\S\ref{sec:img_datasets}) to Wikipedia entries, we can directly map them to Wikidata entities and incorporate them into our taxonomy.\looseness=-1

\subsection{Entity Linked Datasets of Images}
\label{sec:img_datasets}
The iNaturalist21 dataset~\cite{van2021benchmarking} was compiled and curated from the citizen science platform iNaturalist. It contains 2.7 million images spanning 10,000 species. The OVEN dataset~\cite{hu2023open} aggregates several FGVC datasets, linking them to Wikipedia identifiers and natural language questions to support the evaluation of open-domain visual entity recognition. It includes ImageNet21k-P~\cite{ridnik2021imagenetk, ILSVRC15}, iNaturalist2017~\cite{inat17}, Cars196~\cite{cars196}, SUN397~\cite{sun397}, Food101~\cite{food101}, Sports100~\cite{sports100}, Aircraft~\cite{aircraft}, Oxford Flowers~\cite{flowers}, and Google Landmarks v2~\cite{gldv2, ramzi2023optimization}.
We exclude iNaturalist2017 from our OVEN-based evaluations, as we separately analyze the newer iNaturalist21 dataset, which is aligned with the higher-quality Catalogue of Life taxonomy.

\begin{table}
\small
\centering
\setlength{\tabcolsep}{1em}
\begin{tabularx}{\columnwidth}{p{0.1cm}p{1.25cm}|rr|rr}
\toprule
& & \multicolumn{2}{c|}{iNat21} & \multicolumn{2}{c}{OVEN} \\
& Measure & $\tau$-$\hP$ & $\tau$-$\hR$ & $\tau$-$\hP$ & $\tau$-$\hR$ \\
\midrule
& EM & 0.01 & 0.07 & 0.01 & 0.01 \\
& Contained & 0.02 & 0.09 & 0.02 & 0.03 \\
\midrule
\parbox[t]{0.2cm}{\multirow{2}{*}{\rotatebox[origin=c]{90}{string}}} 
& ROUGE & 0.18 & 0.41 & 0.17 & 0.21 \\
& METEOR & 0.18 & 0.40 & 0.18 & 0.22 \\
\midrule
\parbox[t]{0.2cm}{\multirow{3}{*}{\rotatebox[origin=c]{90}{rep.}}}
& BERTscore & 0.01 & 0.31 & 0.27 & 0.18 \\
& SentBERT & 0.02 & 0.25 & 0.34 & 0.31 \\
& NLI & 0.08 & -0.20 & -0.17 & 0.19 \\
\midrule
& CLIP-t2t & 0.14 & 0.46 & 0.25 & 0.15 \\
& \textbf{CLIP-i2t} & \textbf{0.35} & \textbf{0.49} & \textbf{0.35} & \textbf{0.34} \\
\bottomrule
\end{tabularx}
\caption{
\textbf{Correlation of text similarity measures with hierarchical metrics.}
Kendall's $\tau$ correlations between taxonomic metrics $\hP$, $\hR$, and the different similarity measures on the synthetic evaluation sets.
All correlations have a $p$-value $<0.0001$.}
\label{tab:measure_scores}
\end{table}
\section{The Woes of Existing Measures}
\label{sec:evalsimmeasures}

In this section, we evaluate whether non-taxonomic text similarity measures already encode sufficient taxonomic information, e.g., through label overlap or representational similarity. 
To investigate this, we conduct a targeted synthetic experiment comparing the taxonomic similarity measures ($\hP$ and $\hR$), with textual and multimodal similarity measures from \cref{sec:sim_methods} commonly used to assess the quality of language model outputs.
These measures are both string-based and representation-based.

\paragraph{Experimental Setup.}
We construct two synthetic evaluation sets to determine to what extent text similarity measures reflect taxonomic structure. 
We consider the taxonomies from iNaturalist21 and the Wikidata (see \S\ref{sec:datasets}) to generate controlled pairs of labels corresponding to pairs of nodes in each taxonomy. 
A reference node $\refnode$ and a candidate node $\candnode$ are chosen such that $\refnode$ is always a leaf node and $\candnode$ is at some given distance from $\refnode$. 
The distance between two nodes in the taxonomy is the length of the shortest undirected path between the two nodes in the taxonomy. 
To ensure a balanced dataset, we sample the node pairs such that their pairwise distance is uniformly distributed up to a maximum distance of 7, depending on the depth of the node.
To evaluate $\hP$, we sample 100,000 reference--candidate node pairs from the taxonomies.
To evaluate $\hR$ in a controlled fashion, however, we additionally require that $\candnode\in\anc(\refnode)$, i.e.,~that $\hP=1$.
This allows us to isolate the sensitivity of the similarity measures at decreasing levels of specificity.\footnote{
Evaluating $\hP$ while controlling for $\hR$ is only possible for non-leaf reference nodes. Since our datasets always have leaf reference nodes, we did not explore this setting.}
For each pair, we compute both the taxonomy-aware score and text-based measures between the corresponding labels $\reference$ and $\candidate$.
We then measure the rank correlation between the taxonomy-aware score and various text-based measures using Kendall's $\tau$, which evaluates whether the similarity measure reflects taxonomic proximity under controlled conditions. 
By construction, we use the node label $\candidate$ as the prediction to avoid the problem of mapping unconstrained output onto the taxonomy.
All of the measures we consider are text-based (see \cref{app:text_sim_measures} in the supplement for details), except the \mtype{CLIP} image-to-text score, where we sample an image for the reference node from the corresponding dataset and measure \mtype{CLIP} model similarity between the image and the candidate label.
We choose high-performing models that are not prohibitively large for the model-based comparisons.\footnote{All models are downloaded from. \url{huggingface.co}, i.e.,
\mtype{BERTScore} (\modelname{microsoft/deberta-large-mnli}), \mtype{SentenceBERT} (\modelname{sentence-transformers/all-mpnet-base-v2}), \mtype{NLI} (\modelname{MoritzLaurer/mDeBERTa-v3-base-mnli-xnli})
and \mtype{CLIP} (\modelname{apple/DFN5B-CLIP-ViT-H-14}).
}\looseness=-1

\begin{figure}[t]
    \centering
    \includesvg[width=0.95\linewidth]{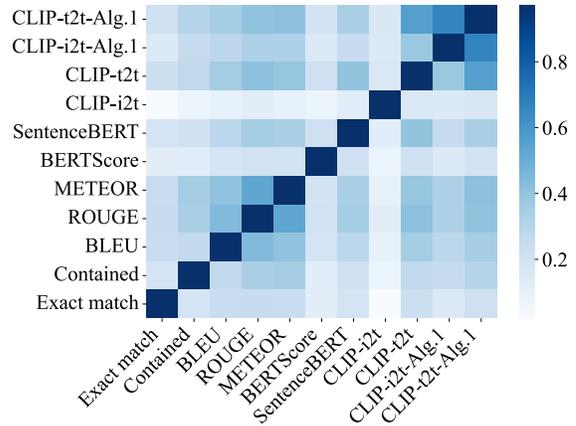}
    \caption{\textbf{Agreement in node placement using different similarity measures.}
    We observe variation in node placement for all measures. Only \mtype{METEOR} and \mtype{ROUGE}, and the \mtype{CLIP} variations, when combined with \cref{alg:tax_mapping} share predictions frequently.\looseness=-1
    }
    \label{fig:heatmap_corr}
    \vspace{-15pt}
\end{figure}

\paragraph{Results.}
\Cref{tab:measure_scores} presents correlations between $\hP$ and $\hR$ on one hand  and the text similarity measures from \cref{sec:sim_methods} and \mtype{CLIP-i2t} on the other,
for the iNaturalist21 and Wikidata labels.
We observe that surface-level measures (\mtype{Exact Match (EM)}, \mtype{Contained}) exhibit a low correlation with the taxonomy-aware measures $\hP$ and $\hR$.
\mtype{ROUGE} and \mtype{METEOR}, which are based on $n$-gram overlap, capture taxonomic similarity more effectively. This is presumably due to label overlap, e.g.,~\concept{gray seal} is a type of \concept{seal}.
The highest correlation is achieved by \mtype{CLIP} image-to-text (\mtype{CLIP-i2t}) for both datasets, followed by \mtype{CLIP} text-to-text (\mtype{CLIP-t2t)}, except for $\hP$ on the iNaturalist21 dataset.
In that case, \mtype{CLIP-t2t} fails to capture deviations from the correct taxonomic path.
This suggests that \mtype{CLIP's} image features may encode more taxonomic information than its text representations, which can struggle due to the low-frequency vocabulary inherent to this task.
\mtype{NLI} is an outlier, exhibiting a negative correlation in most cases.  While NLI captures taxonomic similarity for closely related labels, it performs poorly on more distantly related pairs. This is visualized in \cref{fig:measures_over_hops} in the supplement.\looseness=-1

\paragraph{Summary.}
Overall, we observe that representation-based measures capture some notions of taxonomic distance along the path from the leaf node to the root ($\hR$), but are, for the most part, poorly equipped to measure taxonomic distances across different subtrees ($\hP$).

\begin{table}
\small
\centering
\begin{tabular}{lrrrr}
\toprule
Measure &  $\hP$ & $\hR$ & $\hF$ & Accuracy \\
\midrule
Exact Match & 0.37 & 0.42 & 0.39 & 17.5\% \\
Contained & 0.49 & 0.50 & 0.49 & 24.3\% \\
BLEU & 0.71 & 0.53 & 0.54 & 24.3\% \\
METEOR & 0.62 & 0.65 & 0.63 & 29.1\% \\
ROUGE & 0.62 & 0.66 & 0.63 & 30.0\% \\
BERTScore & 0.45 & 0.52 & 0.48 & 11.3\% \\
SentBERT & 0.58 & 0.67 & 0.62 & 20.0\% \\
CLIP-t2t & 0.72 & 0.79 & 0.75 & 30.3\% \\
CLIP-i2t & 0.58 & 0.71 & 0.23 & 5.3\% \\
CLIP-i2t+\cref{alg:tax_mapping} & \textbf{0.80} & 0.81 & \textbf{0.80} & 44.0\%
\\
CLIP-t2t+\cref{alg:tax_mapping} & 0.79 & \textbf{0.82} & \textbf{0.80} & \textbf{47.1\%}
\\

\bottomrule
\end{tabular}
\caption{\textbf{Evaluating taxonomy mapping with human annotation.} We report the performance of text similarity measures (directly) and CLIP image-to-text similarity for mapping VLM text predictions to the taxonomy using a subset of iNaturalist21 samples that were manually mapped to the correct nodes.}
\label{tab:positionmetrics}
\vspace{-15pt}
\end{table}

\section{Mapping VLM Predictions to Taxonomies}
\label{sec:mapping}
The key challenge for taxonomy-aware evaluation of unconstrained text generated from VLMs is mapping the text generated by a VLM onto nodes in a given taxonomy.
A baseline approach to this mapping is to use text similarity measures (e.g.,~those in \cref{sec:sim_methods}) to calculate the best match between the generated text and the labeled nodes in a taxonomy.
We also develop a heuristic approach that performs this mapping using a general similarity measure as a subroutine, which makes using \mtype{CLIP text-to-image} similarity feasible.
Specifically, this algorithm combines string-based matching with scores from a given similarity measure to position predictions onto the taxonomy.
We describe this procedure in words below and give pseudocode in \cref{sec:algo} (\mtype{\cref{alg:tax_mapping}} supplement).\looseness=-1

Our goal is to map each VLM-generated prediction string to a node in the taxonomy using a series of increasingly flexible matching strategies. Generally, we attempt exact matches first, then allow for partial overlaps or ambiguous cases, always preferring more specific nodes when multiple candidates are viable.
Before matching, we normalize all predictions and node labels by lowercasing, replacing dashes with spaces, and stripping other punctuation.
We then compute similarity scores between the prediction and all node labels using the \mtype{CLIP} similarity measures.
More specifically, the mapping proceeds in stages.
We first check whether any of the top-$k$ ($k=10$) highest-scoring nodes, according to the \mtype{CLIP} measure applied to their labels, appear verbatim in the prediction string; if so, we return the corresponding node. 
If this fails, we search for verbatim matches against all other labels in the taxonomy. If no full-string matches are found, we repeat these checks using $n$-gram overlap for $n \in \{4, 3, 2\}$.
If multiple candidates remain with similar scores, we apply a fallback based on score ambiguity: if the difference between the top two scores, and between the top and $k^\text{th}$ score, are both below threshold values, we treat the case as ambiguous.
We then search for a common ancestor among the top candidates.
If any node appears sufficiently often (at least four times) in the ancestor sets of multiple nodes in the top-$k$ highest-scoring nodes we return the most specific such node, defined as the one deepest in the taxonomy.
If none of the prior criteria are met, we fall back to returning the single highest-scoring candidate.
Thus, the algorithm always returns a node in the taxonomy.
If multiple nodes match the above criteria, we return the most specific one, i.e., the node deepest in the taxonomy.
The ``shared ancestor'' step is particularly beneficial for the CLIP image-to-text similarities since we do not have images corresponding to higher-order taxonomic concepts.\looseness=-1

\subsection{Mapping Quality}
To evaluate the quality of the taxonomic node proposals, we measure their agreement against manually annotated node positions. 
We compare the similarity measures described in \cref{sec:sim_methods} (and Supplement \cref{app:text_sim_measures}) directly against all candidate node labels.
We then do the same using \cref{alg:tax_mapping}, combined with the \mtype{CLIP} measures.
To construct the test set, we gather 416 unique VLM outputs for the iNaturalist21 dataset, randomly sampled from the generations described in \S\ref{sec:comparing_vlms}. 
For each output, we manually find the best matching taxonomic node given the string, e.g., matching \modelpred{This is a bird} to \concept{Aves/Birds}.\looseness=-1

\paragraph{Results.} 
Results are shown in \cref{tab:positionmetrics}.
We find string-matching measures (\mtype{Exact Match} and \mtype{Contained}) do fairly poorly, while other surface-level measures (\mtype{BLEU}, \mtype{METEOR}, and \mtype{ROUGE}) perform slightly better. The representation-based approaches do not perform better in general, except for \mtype{CLIP text-to-text}.
\cref{fig:heatmap_corr} shows that the string-level measures have considerable diversity in their predictions, while the representation-based measures show more diversity.
The best performance is found with \mtype{CLIP text-to-text} and \mtype{CLIP image-to-text} using the algorithm described above (\cref{alg:tax_mapping}), which results in the same $\hF$ score but slightly improved exact node match accuracy score for \mtype{CLIP text-to-text + \cref{alg:tax_mapping}}.

\begin{table}
\fontsize{10}{10}\selectfont
\centering
\renewcommand{\arraystretch}{1.4} % vertical padding
\setlength{\tabcolsep}{0.35em} % for the horizontal
\resizebox{\columnwidth}{!}{%
\begin{tabular}{@{}lrrc@{}}
\toprule
Model Name            & \# Parameters & Training Pipeline   & OpenVLM Avg.\\ \midrule
GPT-4 \cite{openai_gpt-4_2023}                & Unknown       & Unknown                               & 63.1                     \\
ILmXC2 \cite{dong2024internlmxcomposer2} & 7B            & Pretr., instr.-tuning       & 62.2                     \\
OLM12B \cite{yu2023rlhfv}            & 12B           & Pretr., instr.-tuning, RLHF & 54.6                     \\
QVLChat \cite{Qwen-VL}           & 9.6B          & Pretr., instr.-tuning       & 51.3                     \\
OLM3B \cite{minicpm2024}  & 3B            & Pretr., instr.-tuning       & 47.8                     \\
LLaVA \cite{liu2023improvedllava}        & 7.2B          & Pretr., instr.-tuning       & 45.9                     \\
QwenVL \cite{Qwen-VL}                & 9.6B          & Pretr.                           & 21.7                     \\
Fuyu \cite{fuyu-8b}                 & 8B            & Pretr.                           & N/A                  \\
\bottomrule

\end{tabular}
}
\caption{\textbf{Overview of evaluated VLMs.} Details on the models we evaluate, including number of parameters, training method, and average performance on a large vision--language benchmark.}
\label{tab:models}
\end{table}

\section{Evaluating and Comparing VLMs}
\label{sec:comparing_vlms}
We compare one closed model and seven publicly available VLMs on fine-grained visual classification using our taxonomy-aware evaluation scheme.
To map (unconstrained) strings generated from a VLM to nodes in the taxonomy, we use the best-performing similarity method (\mtype{CLIP-t2t} + \cref{alg:tax_mapping}), enabling the computation of $\hP$, $\hR$, and $\hF$.
These scores offer insight into the level of accuracy ($\hP$) and specificity ($\hR$) of VLM predictions for a given taxonomy.
The resulting rankings differ significantly from those based on standard text similarity measures, highlighting the taxonomy-aware measure's ability to capture previously unmeasured aspects of model performance.\looseness=-1

\paragraph{Experimental Setup.}
We compare eight VLMs that differ in model size, training configuration (pretrained-only, instruction-tuned, and preference-aligned~\cite{ouyang2022training}), and performance on the Open VLM Leaderboard~\cite{2023opencompass}, a collection of 13 VL benchmarks with 38 models.
We summarize model sizes, training configuration, and average performance on the Open VLM leaderboard in \Cref{tab:models}.
For both the OVEN-sourced FGVC datasets and iNaturalist21, we randomly sample 6000 images from the validation set.
For iNaturalist21, we use the prompt \modelpred{What species is this?}, while for OVEN the question is included in the dataset and varies by object type.
To understand how well we can control $\hR$ and $\hP$ for instruction-tuned models, we construct two different sets of instructions: (1)~one containing a minimal instruction for answering in a specified format, and (2)~one containing an instruction to be specific.
We use slight variations of the prompt formatting to fit each model (as specified by the model developers);
the exact prompts are provided in the supplement (\cref{tab:sup_prompts2} and \cref{tab:sup_prompts3}).
Because non-instruction-tuned models do not respond meaningfully to instructions, we prompt them with a straightforward question--answer template \modelpred{Q:~\predvar{\{\}} A:}. We use an orange-colored \predvar{\{\}} to indicate a variable in the template.\looseness=-1

\subsection{Experimental Results}  
\label{sec:expresults}

\paragraph{How does the ranking of VLM performance change when considering taxonomy-aware measures?}

\begin{figure}
    \centering
    \includegraphics[width=\columnwidth]{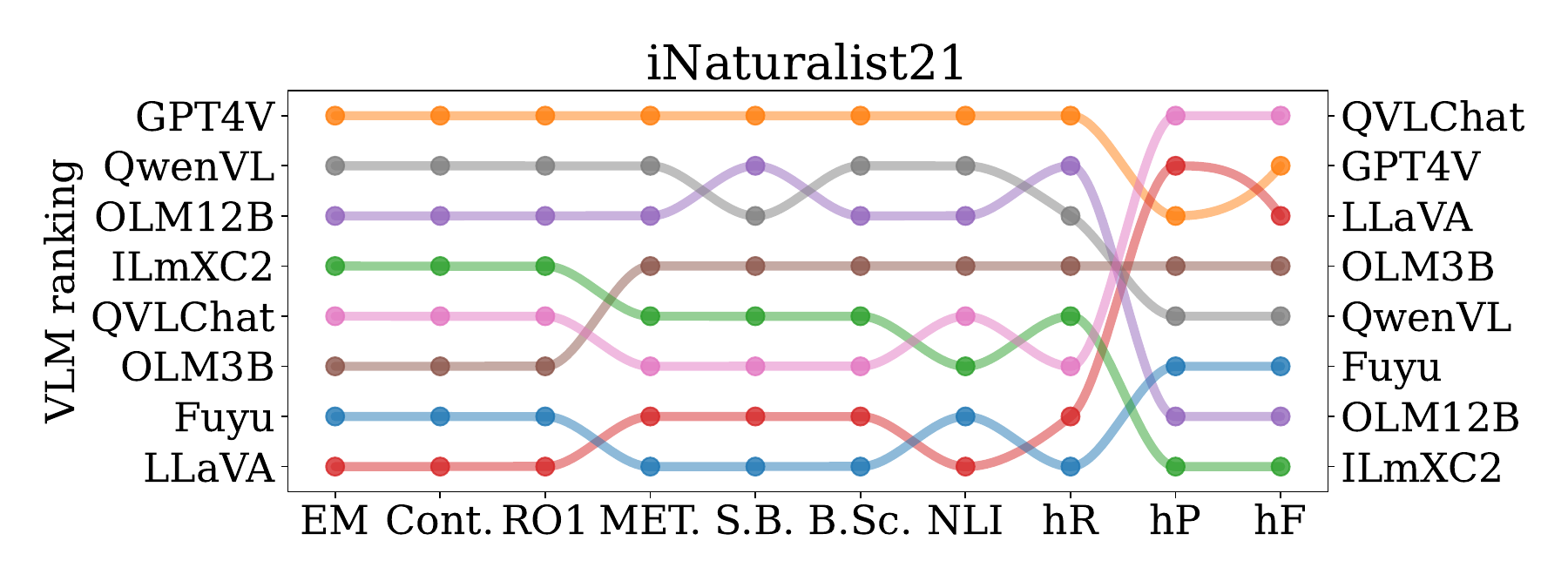}
     \includegraphics[width=\columnwidth]{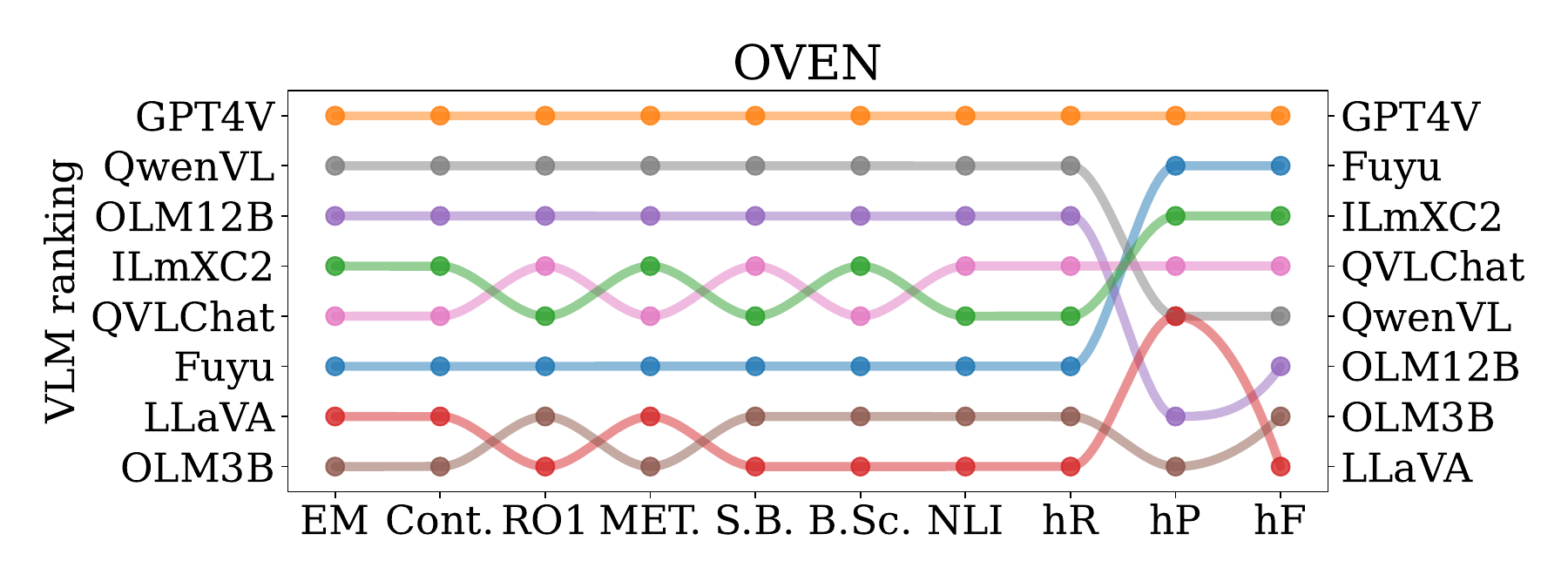}
    \caption{\textbf{Ranking of VLMs for the iNaturalist21 (top) and OVEN (bottom) datasets.} We evaluate the ranking of VLMs (vertical axis) based on different evaluation measures (horizontal axis). The best model is shown in the top row. On the left, we see the model names ranked by \mtype{exact match} and on the right ranked by the hierarchical F1 $\hF$. See \cref{app:ranking}, \cref{tab:ranking-values} in supplement for exact numbers.}
    \label{fig:ranking}
\end{figure}

In \Cref{fig:ranking}, we rank eight VLMs based on their performance on FGVC according to the different similarity measures corresponding to the values given in \cref{tab:ranking-values} and \cref{tab:ranking-values-inat}  (\cref{app:ranking} in supplement). 
We find that several text similarity measures yield similar rankings when used as direct evaluation criteria. In contrast, the taxonomy-aware measures lead to notable changes in model rankings. For instance, LLaVA ranks lowest under \mtype{Exact Match} for iNaturalist21, yet achieves high hierarchical precision ($\hP$) and, consequently, a strong hierarchical F1 ($\hF$). This indicates that LLaVA deviates from the correct taxonomy path less frequently than other models. However, its low hierarchical recall ($\hR$) suggests that its predictions are often not specific.
Another example is GPT-4, which ranks highest on all measures except $\hP$ and $\hF$ for iNaturalist21. 
Compared to QVLChat, which is ranked first in both $\hP$ and $\hF$, we observe that GPT-4 makes more specific but erroneous predictions.
Our taxonomy-aware measure offers a complementary perspective to traditional metrics by revealing performance differences that standard similarity-based measures overlook.

\begin{figure}
    \centering
    \includegraphics[width=\columnwidth]{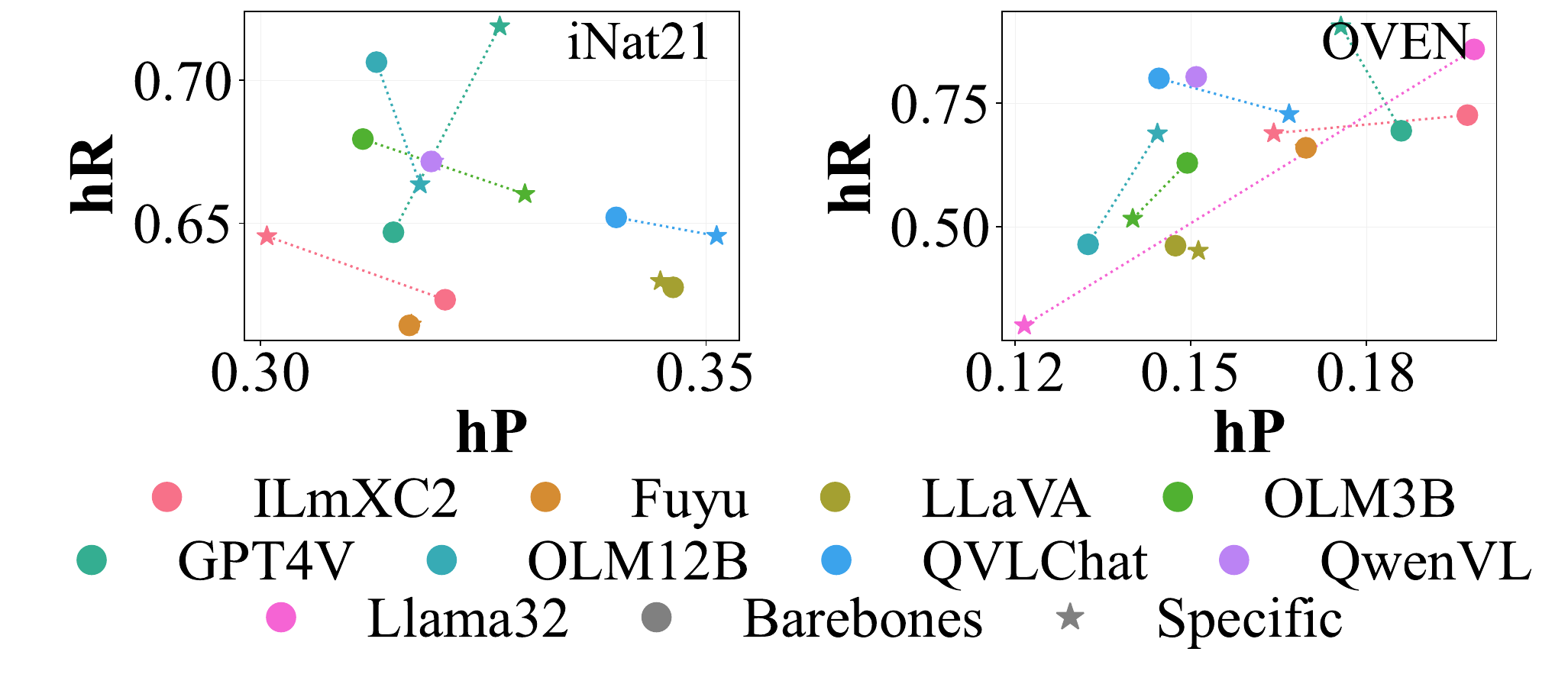}
   \caption{\textbf{Prompt sensitivity captured by $\hR$ and $\hP$.} 
   We compare two prompt versions, a minimal barebones version and one that asks for a taxonomically specific prediction.}
    \label{fig:acc_spec_vlm}
\end{figure}

\paragraph{Do taxonomy-aware measures capture prompt sensitivity?}
We prompt the VLMs using two distinct prompts, as described earlier, to examine how their predictions shift in terms of hierarchical precision ($\hP$) and recall ($\hR$), shown jointly in \Cref{fig:acc_spec_vlm}.
Fuyu and QwenVL are not instruction-tuned and therefore do not respond differently, while LLaVA shows only minimal changes. In contrast, the other models clearly adjust their predictions based on the different instructions in the prompts.
On the iNaturalist21 dataset, all models except GPT-4 exhibit a trade-off between $\hP$ and $\hR$: higher precision tends to come at the cost of recall. 
GPT-4, however, offers both more accurate and more fine-grained classifications when prompted appropriately.
On the OVEN dataset, results are more mixed: roughly half of the models improve on both axes with prompt changes.
These findings suggest that each application may prioritize different target values for $\hP$ and $\hR$, and that models can adapt substantially to meet those needs. In such cases, prompt design becomes a powerful paradigm for tuning model behavior as further explored in the next section.

\subsection{An Application: High-Precision Classification}
\label{sec:birdclass}
\begin{figure}
    \centering
    \includesvg[width=0.88\columnwidth]{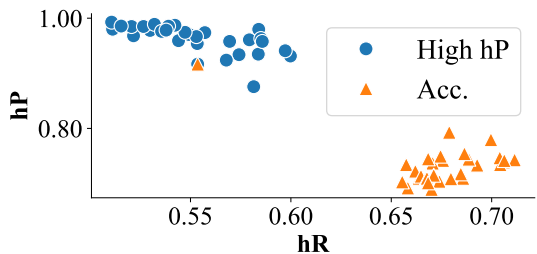}
    \caption{\textbf{Prompt tuning results for bird classifier.} For \emph{High hP} ({\textcolor{mplblue}{$\bullet$}}),
 we target prompts that prioritize $\hP$. For \emph{Acc} ({\textcolor{mplyellow}{$\blacktriangle$}}) we optimize for prompts that give higher binary accuracy. This demonstrates how $\hP$ and $\hR$ can help tune a VLM application.}
    \label{fig:application}
\end{figure}

To demonstrate the usefulness of $\hR$ and $\hP$ in tuning prompts for a specific application, we prompt Llama 3.2 Vision (11B) \cite{touvron2024llama3} to classify birds.
The application goal is to minimize the amount of wrong information in the prediction,
which is equivalent to maximizing hierarchical precision but hard to formulate in terms of exact match binary classification accuracy.
We use prompt-tuning with $\hP$ and $\hR$ as feedback signals, and compare to the same procedure but with accuracy as the feedback signal.
We start with a basic prompt \modelpred{What is this an image of? Answer in the format `A: \textless answer\textgreater.'} and task ChatGPT \cite{openai_gpt-4_2023} to improve on the prompt for 30 iterations. 
A system prompt describes the scores and the goal of developing a conservative classifier that gives a less specific prediction if uncertain (see \cref{app:birdclass} in the supplement for more details).
In each iteration, we include past prompts along with their resulting $\hR$ and $\hP$ scores, or the accuracy score, in a few-shot learning setup.
We tune the prompts with 20 images and evaluate on 200 images of birds from iNaturalist21.
The results are shown in \cref{fig:application}. While this is a toy example, we see how the taxonomy-aware measures not only enable a higher granularity of model performance analysis, but can also facilitate targeted application development.\looseness=-1

\section{Discussion and Limitations}
Evaluating unconstrained text generated from a VLM is inherently challenging: there are countless entities to describe and even more ways to describe them.
As the computer vision community shifts from closed-set recognition to open-ended recognition, evaluation increasingly resembles an NLP task.
In this paper, we introduce structure into VLM evaluation by leveraging curated taxonomies and established hierarchical evaluation measures. 
Despite this, several limitations remain.
First, placing unconstrained text on a taxonomy is a low-resource problem; there are no large-scale datasets available to evaluate our mapping approach, nor sufficient data to train a dedicated classifier that links VLM outputs to taxonomy nodes.
To address this, we repurpose existing text similarity measures, along with CLIP, to perform the mapping. While our evaluation shows that this approach is non-trivial and introduces errors that affect downstream VLM evaluation, we are optimistic about future improvements, such as training on synthetic data generated by language models.
Second, taxonomies, especially those derived from Wikidata, are inherently noisy. 
Even expert-curated taxonomies like the one used for iNaturalist21 can have uneven subtree granularity, complicating interpretations based on globally averaged measures.
One potential solution would be to employ expert input to assign weighted edges to reflect semantic distances between nodes.\looseness=-1

\section{Conclusion}
Text similarity measures from NLP are a natural place to look for evaluation of unconstrained VLM output. 
Additionally, many FGVC tasks come with side information about the class structure, namely taxonomies, which text similarity measures do not fully capture, as we have shown in this paper.
To solve this gap, we have proposed a framework for using hierarchical evaluation measures to evaluate VLMs for FGVC, enabling a more fine-grained view of the specificity and accuracy of these models.\looseness=-1

\section*{Acknowledgements}
Vésteinn Snæbjarnarson, Serge Belongie, Nico Lang, and Stella Frank are supported by the Pioneer Centre for AI, DNRF grant number P1. Niklas Stoehr is supported by the Swiss Data Science Center (SDSC) PhD Fellowship.

{
    \small
    \bibliographystyle{ieeenat_fullname}
    \bibliography{main}
}

\setcounter{page}{1}
\clearpage

\maketitlesupplementary

\section{Elaborating on Text Similarity Measures}
\label{app:text_sim_measures}
In this section, we elaborate on the different text similarity measures used in this paper and discuss trade-offs across the measures.
The methods are ordered roughly in terms of computational overhead, with string-matching methods first (\mtype{Exact Match}, \mtype{Contained}), then methods based on $n$-grams (\mtype{ROUGE}, \mtype{METEOR}), and finally methods that use forward passes through pre-trained large neural networks (\mtype{BERTScore}, \mtype{SentenceBERT}, \mtype{NLI}, \mtype{CLIP}).
\begin{itemize}
\item \textbf{\mtype{Exact Match (EM)}} measures whether a model's predicted output exactly matches the label. This measure is commonly used in, e.g., question-answering systems. 
\item \textbf{\mtype{Contained}}, like \mtype{EM}, measures a model's prediction against a gold label by lexical matching. However, this measure is looser, i.e., as long as the prediction contains the label as an exact subsequence, it will be considered correct. Such a measure will have higher recall but lower precision than EM, since it may produce more false positives as well as fewer false negatives; as such, this is particularly susceptible to false positives in predictions which contain the gold label but mean something different in the text context \citep{wang2023evaluating}. 
    In the iNaturalist21 setting, Contained will often match higher-level labels 
    (e.g., the species label \concept{chestnut-collared swift} contains the genus label \concept{collared swift} which contains the family label \concept{swift}.)

\item \textbf{\mtype{BLEU}} \citep{bleu} is an n-gram overlap measure used originally for evaluating machine translation.
It measures how well a predicted string matches a reference (label) string, based on the matched $n$-grams between the two.
\mtype{BLEU} is precision-oriented: it captures the proportion of $n$-gram spans in the predicted string that also appear in the reference string.
Predictions that mostly coincide with the reference string with little extra text will score higher.
We use \mtype{BLEU2} ($n=2$, i.e.,~bigrams) with smoothing.
\item \textbf{\mtype{ROUGE}} \citep{lin-2004-rouge} is a traditional measure used in automatic summarization.
Like \mtype{BLEU}, \mtype{ROUGE} is based on n-gram overlap, but unlike \mtype{BLEU},
\mtype{ROUGE} is a recall-oriented measure: if many $n$-grams contained in the reference string are contained in the predicted string, then the prediction will have a high \mtype{ROUGE} score regardless of whether it contains many other irrelevant $n$-grams.
It will thus have a similar pattern of behavior as Contained.
We use \mtype{ROUGE1}, i.e.,~measuring unigram recall.
\item \textbf{\mtype{METEOR}} \citep{banerjee-lavie-2005-meteor} scores a model's prediction against a gold label based on two factors: first, the unigram precision-recall harmonic mean, i.e., the harmonic (oftentimes weighted) average between the precision and recall of the prediction and label at the unigram level, and an alignment penalty, which intuitively aims to capture how close the ordering of the words in the prediction match the ordering of the words in the label. This alignment penalty is based on the number of consecutive unigram chunks that can be aligned between the prediction and label (where fewer chunks means a lower penalty). It also flexibly matches word stems, paraphrases, or synonyms, if unigrams do not match exactly. This measure captures tends to correlate better with human judgments than other lexical matching measures like \mtype{ROUGE}. 
\item \textbf{\mtype{BERTScore}} \citep{bert-score} is a representation-based measure which compares the prediction and label based on semantic similarity. To compute the BERTScore of a prediction against a label, one must first compute the token-level \mtype{BERT} representations \citep{devlin_bert_2019} for each token in the prediction and label. Subsequently, the semantic similarity of each token-pair between prediction and label is computed using cosine similarity. These token-level cosine similarities are then aggregated to compute the precision, recall, and F1. By relying on similarity in contextual token representations, \mtype{BERTScore} is better at capturing paraphrases than the above measures based on lexical matching. It is also better at settings in which semantic similarity is an important criterion to judging the prediction and label, but not the exact tokens.
\item \textbf{\mtype{SentenceBERT}} \citep{reimers_sentence-bert_2019} is also a representation-based measure to compare the prediction and label based on semantic similarity. However, \mtype{SentenceBERT} aggregates the representations for a sentence into a single fixed-length representation representing the full sentence. Thus, in comparing the prediction and label, one computes the cosine similarity between the \mtype{SentenceBERT} representation for both prediction and label.
\item \textbf{\mtype{NLI}} \citep{bowman_large_2015, wang_entailment_2021, lefebvre_rethinking_2023, xie2019visual} uses textual entailment to judge the specificity of a prediction. If the prediction entails the label, then the prediction is more specific than the label. If the label entails the prediction, then the label is more specific. If both entail each other, this suggests the prediction and label are perfect matches. If neither entails the other, then this suggests low hierarchical precision. 
\item \textbf{\mtype{CLIP text-to-text}} \citep{clip} compares the prediction to the label by comparing the cosine similarity of the \mtype{CLIP} text representation of the predicted string and label string. 
\item \textbf{\mtype{CLIP image-to-text}} \citep{clip} compares the prediction to an image sampled from the label category by comparing the cosine similarity of the \mtype{CLIP} text representation of the predicted string and the \mtype{CLIP} image representation of the image matching that label. 
Since \mtype{CLIP} is explicitly optimized for matching images and text representations, we hypothesized \mtype{CLIP} may be better suited for comparing hierarchical similarity between text (prediction) and image (corresponding to label). \looseness=-1
\end{itemize}

\section{Mapping Predictions Onto a Taxonomy}
\label{sec:algo}

\newcommand{\thresholdsim}{\mymacro{\tau_{\mathrm{\texttt{topk}}}}}
\newcommand{\thresholdsimtop}{\mymacro{\tau_{\mathrm{\texttt{top2}}}}}
\newcommand{\thresholdvote}{\mymacro{\tau_{\mathrm{\texttt{vote}}}}}
\newcommand{\ngram}{\mymacro{\mathrm{ngr}}}
\newcommand{\topk}{\mymacro{S_k}}
\newcommand{\topknodes}{\mymacro{\nodes_k}}
\newcommand{\nodelab}{\mymacro{\node_{\text{lab}}}}

The algorithm we use for mapping predictions onto the taxonomies is given in \cref{alg:tax_mapping}. It is described in section \cref{sec:mapping} in the main paper. We use the parameters \texttt{k=10, thr\_topk=0.0015, thr\_top2=0.001, thr\_vote=4}.

\begin{algorithm}
    \caption{Mapping predictions onto a taxonomy}
\begin{minted}[fontsize=\scriptsize]{python}

def anc(node):
  ancs = [node]
  par = node.parent
  while par:
    ancs.append(par)
    par = par.parent
  return ancs

def n_gram(text, n):
  spl = text.split()
  return [" ".join(spl[i:i+n]) for i in range(len(spl)-n+1)]

def map_tax(pred, T, m, k, thr_topk, thr_top2, thr_vote):
  """
  Map a prediction to the most similar node in a taxonomy.
  *Inputs*
  pred: str, model prediction
  T: object, taxonomy tree relating the nodes
  m: function, similarity measure
  k: int, number of top similar nodes to consider
  thr_topk: float, max difference between top-1 and top-k
  thr_top2: float, max difference between top-1 and top-2
  thr_vote: int, min number of votes for node to be selected   
  """
  S = [(m(pred, v.label), v) for v in T]  # similarity 
  S.sort(key=lambda x: -x[0])  # sort nodes by similarity
  S, V = zip(*S)  # store sorted similarity S and nodes V
  S_k = softmax(S[:k])  # normalize top-k similarity scores
  
  # Contains check: 
  # return most specific node where pred. contains the label
  cand = None
  for v in V:
    if v.label in pred:
      if cand is None
          or len(anc(v)) > len(anc(cand)):
        cand = v  # store most specific node
    if cand is not None and V.index(v) == k - 1:
       return cand  # We found a hit in top-k
  if cand is not None:
    return cand

  # n-gram check: 
  # return most specific node with overlapping n-grams
  for n in (4, 3, 2):
    cand = None
    pred_ngrams = n_gram(pred, n)
    for v in V:
      v_ngrams = n_gram(v.label, n)
      if pred_ngrams.intersect(v_ngrams):
        if cand is None 
            or len(anc(v)) > len(anc(cand)):
          cand = v  # store most specific node
      if cand is not None and V.index(v) == k - 1:
        return cand  # We found a hit in top-k
    if cand is not None:
      return cand
  
  # Voting: 
  # return most specific common ancestor in top-k nodes
  # with minimum thr_vote number of occurrences, if multiple
  # such nodes, choose the most frequent one.
  if S_k[0] - S_k[1] < thr_top2 
    and S_k[0] - S_k[-1] < thr_topk:
    votes = defaultdict(lambda: defaultdict(int))
    for v in V[:k]:
      for i, node in enumerate(anc(v)):
        votes[i][node] += 1
    for i, counts
        in sorted(votes.items(), key=lambda x: -x[0]):
      node, count = max(counts.items(), key=lambda x: x[1])
      if count > thr_vote: return node

  # Fallback: return most similar node
  return V[0]
  
\end{minted}
\label{alg:tax_mapping}
\end{algorithm}

Both Wikidata and the iNaturalist21
taxonomies come with canonical and alternative label names. We do not make use of the alternative non-English Wikidata labels.
For the iNaturalist21 dataset, we make use of both the canonical Linnean Latin names, and the English common names.
In \cref{alg:tax_mapping}, we compare predictions both to \concept{bird} and Latin \concept{Aves}.
The text similarity measures use the common (English) name whenever it is available.

For the direct comparison in \cref{alg:tax_mapping} we do basic tokenization: we strip punctuation, whitespace, and lowercase. In the case of \mtype{Exact Match}, \mtype{Contained}, \mtype{ROUGE}, and \mtype{BLEU} we also stem the words (e.g..~removing the difference between singular and plurals).

\section{Visualization of Correlations}
\begin{figure*}[htb!]
    \centering   
    \begin{subfigure}[b]{\textwidth}
        \includegraphics[width=0.475\textwidth]
    {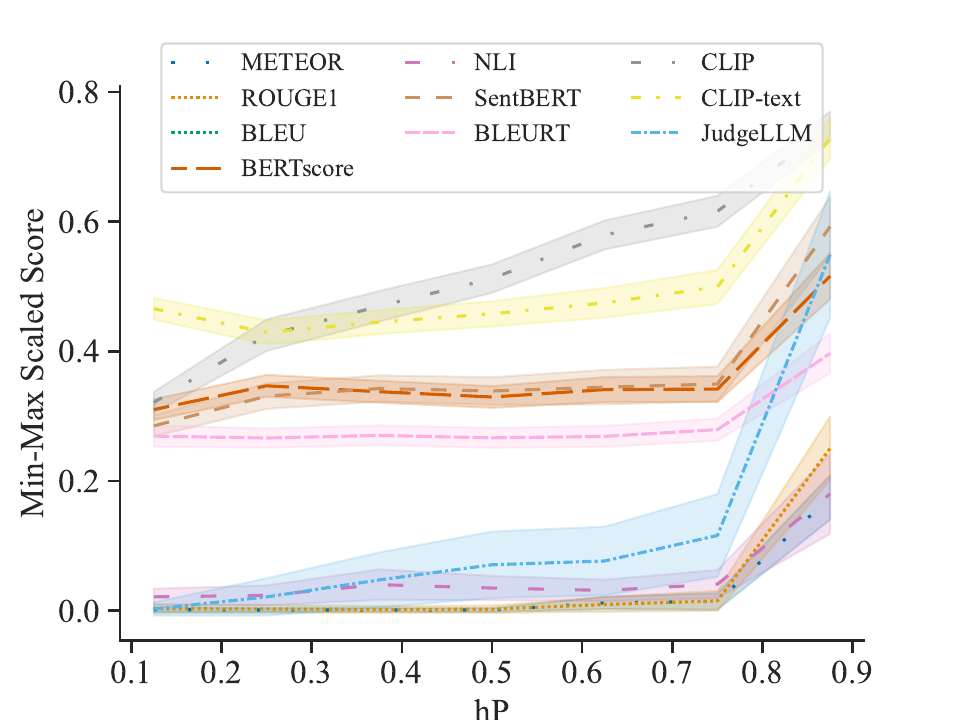}
    \hfill
   \includegraphics[width=0.475\textwidth]{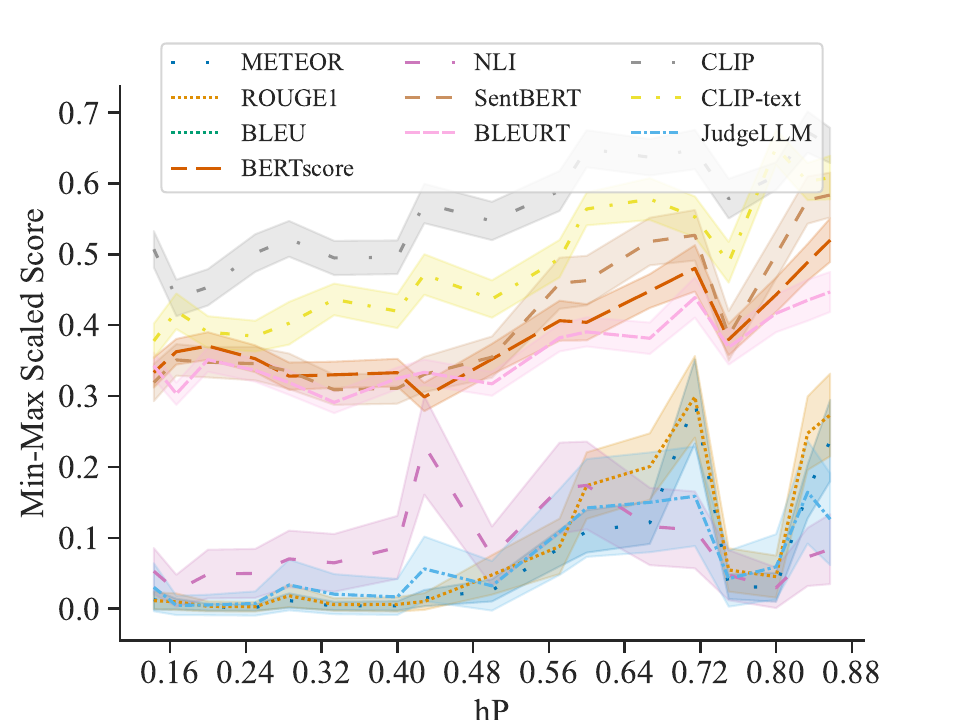}
    \caption{Similarity measures plotted against $\hP$.}
    \label{subfig:accM}
    \end{subfigure}
    \begin{subfigure}[b]{\textwidth}
        \includegraphics[width=0.475\textwidth]
        {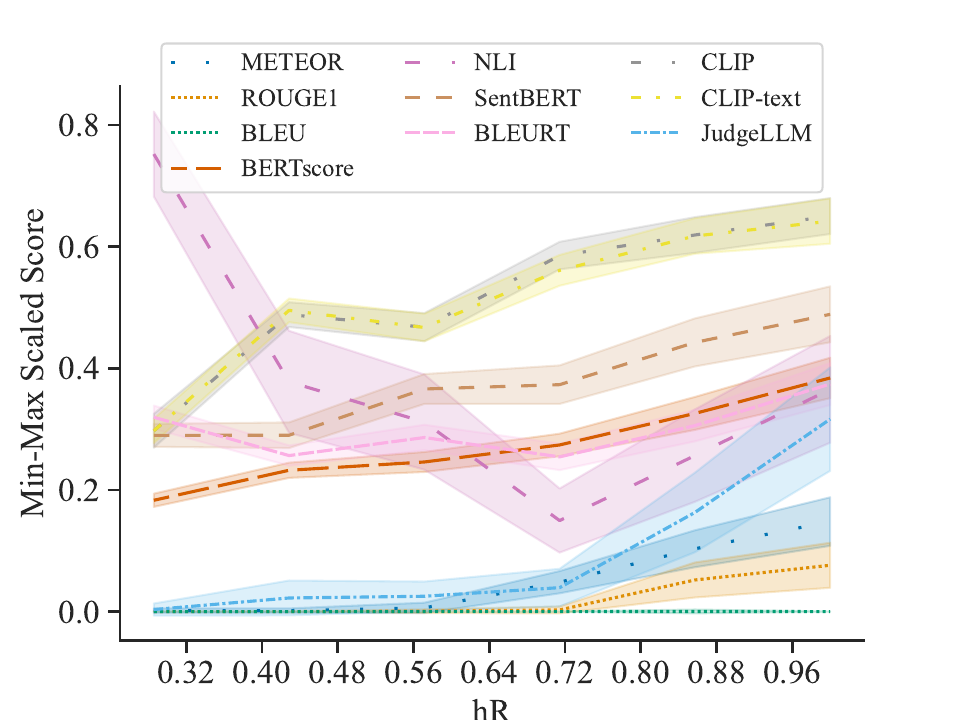}
        \hfill
       \includegraphics[width=0.475\textwidth]{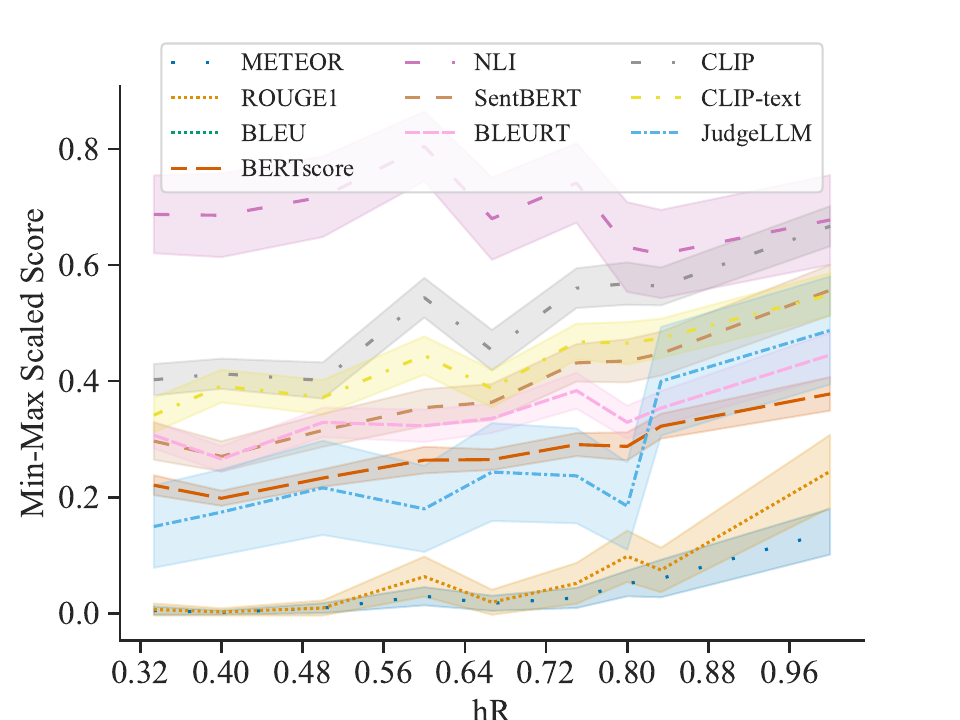}
        \caption{Similarity measures plotted against $\hR$. }
    \end{subfigure}
    \caption{\textbf{Text similarity measures vs. taxonomy-aware measures.} iNaturalist21 (left) and Wikidata (right) measures as a function of $\hP$ (top) and $\hR$ (bottom). Error bars are standard deviation scaled by $\frac{1}{5}$ to improve readability while allowing for a relative comparison of the measures.}
    \label{fig:measures_over_hops}
\end{figure*}

The data used to compute the correlations reported in \cref{tab:measure_scores} are visualized in \cref{fig:measures_over_hops}.

\section{Prompts Used for the VLMs}
The full prompts used for the VLM generations are given in \cref{tab:sup_prompts2} and \cref{tab:sup_prompts3}. The prompts vary slightly due to differences in how the models are prompted. For iNat21, the question is always: \modelpred{What species is this?}, while for OVEN the question varies by object type, e.g., \modelpred{What is the model of this aircraft?}

\section{Ranking of VLMs}
\label{app:ranking}
We plot the ranking of different VLMs in \cref{sec:expresults}, the numbers behind the plot are given in \cref{tab:ranking-values} (OVEN) and \cref{tab:ranking-values-inat} (iNaturalist21).

\begin{table*}[htb]
\centering
\begin{tabularx}{\textwidth}{lcccccccccc}
\toprule
Model & \mtype{EM} & \mtype{Contained} & \mtype{ROUGE1} & \mtype{METEOR} & \mtype{SentBERT} & \mtype{BERTScore} & \mtype{NLI} & $\hR$ & $\hP$ & $\hF$ \\
\midrule
Fuyu      & 23.1  & 23.1  & 43.2 & 29.0  & 55.8 & 57.4 & 76.1 & 0.660 & 0.166 & 0.265 \\
GPT4V     & 78.9  & 78.9  & 81.0 & 70.4  & 85.0 & 87.1 & 88.6 & 0.876 & 0.168 & 0.282 \\
OLM12B    & 44.5  & 44.5  & 59.3 & 43.2  & 67.8 & 69.1 & 81.9 & 0.727 & 0.145 & 0.242 \\
QwenVL    & 50.2  & 50.2  & 65.1 & 48.5  & 73.2 & 73.1 & 86.4 & 0.804 & 0.151 & 0.254 \\
LLaVA     & 18.2  & 18.2  & 25.9 & 19.5  & 41.7 & 50.0 & 51.3 & 0.445 & 0.151 & 0.225 \\
OLM3B     & 14.0  & 14.0  & 31.9 & 19.3  & 46.2 & 50.8 & 64.8 & 0.538 & 0.144 & 0.227 \\
QVLChat   & 34.4  & 34.4  & 52.5 & 35.5  & 62.7 & 63.3 & 80.6 & 0.707 & 0.159 & 0.260 \\
ILmXC2    & 36.8  & 36.8  & 50.6 & 38.4  & 61.5 & 64.1 & 76.7 & 0.687 & 0.161 & 0.261 \\
\bottomrule
\end{tabularx}
\caption{\textbf{VLM evaluation for the OVEN dataset.} Mean similarity measure results using the minimal templates for each model.}
\label{tab:ranking-values}
\vspace{2em}
\centering
\begin{tabularx}{\textwidth}{lcccccccccc}
\toprule
Model & \mtype{EM} & \mtype{Contained} & \mtype{ROUGE1} & \mtype{METEOR} & \mtype{SentBERT} & \mtype{BERTScore} & \mtype{NLI} & $\hR$ & $\hP$ & $\hF$ \\
\midrule
GPT4V     & 14.4  & 14.5  & 15.7  & 18.9  & 47.6  & 54.6  & 37.6  & 0.687 & 0.336 & 0.451 \\
LLaVA     & 0.409 & 0.409 & 1.28  & 5.19  & 36.9  & 45.8  & 20.3  & 0.630 & 0.346 & 0.447 \\
OLM12B    & 4.07  & 4.10  & 5.28  & 10.9  & 42.5  & 50.2  & 29.2  & 0.680 & 0.300 & 0.416 \\
Fuyu      & 0.577 & 0.577 & 1.39  & 4.91  & 36.2  & 45.5  & 22.1  & 0.615 & 0.316 & 0.417 \\
QwenVL    & 5.49  & 5.56  & 7.08  & 11.2  & 42.3  & 50.5  & 33.2  & 0.672 & 0.319 & 0.433 \\
OLM3B     & 0.882 & 0.882 & 1.95  & 7.69  & 40.0  & 47.6  & 24.1  & 0.666 & 0.325 & 0.437 \\
ILmXC2    & 1.34  & 1.34  & 2.44  & 7.60  & 39.2  & 47.4  & 22.9  & 0.654 & 0.296 & 0.408 \\
QVLChat   & 1.15  & 1.15  & 2.23  & 6.32  & 38.1  & 46.3  & 23.2  & 0.635 & 0.357 & 0.457 \\
\bottomrule
\end{tabularx}
\caption{\textbf{VLM evaluation for the iNaturalist21 dataset.} Mean similarity measure results using the minimal templates for each model.}
\label{tab:ranking-values-inat}
\end{table*}

\section{Bird Classifier Example}
\label{app:birdclass}
\begin{table*}[t]
\small
\centering
\begin{tabularx}{\textwidth}{>{\raggedright\arraybackslash}m{2.5cm} >{\raggedright\arraybackslash}X}
\toprule
Approach & Template \\
\midrule
$\hR/\hP$ aware & \modelpred{You are an AI assistant helping to generate effective prompt templates for vision-language models to identify birds in images. The prompt templates should help the model provide accurate species identification that don't contain wrong information, if in doubt we want to back off and ensure the hP value stays as close to 1 as possible. Analyze the previous prompts and their performance metrics (hR: Hierarchical Recall, hP: Hierarchical Precision) to generate an improved template which produces as high scores for both hR and hP as possible. The goal is to make the model we are tuning never return any false information -- this is measured by the hP. Then we also want as much correct information as possible, this is the hR. hP=1 is a priority. A simple baseline for this would be to always answer ``Bird''. But this would always give 0.5 for hR which is as low as we can get, so really try and make this higher, while still aiming for hP=1. Note that the target model is a small Llama 3.2 vision, so it will be sensitive to variations in the prompt. Here's an explanation of the two metrics. Given a taxonomy of entities: * Hierarchical Precision (hP) captures the amount of incorrect information in the prediction, i.e., deviations from the correct path, relative to the extent of the shared, correct, path. * Hierarchical Recall (hR) measures the amount of correct information in the prediction: how much the predicted node's path, to the root, intersects with the correct path. It thus penalizes missing coverage of the target path, especially early deviations. This can be seen as measuring the specificity of the prediction, in particular when the hP is 1. * Both HP and HR achieve the highest score (1.0) when the target and predicted paths coincide. Less-than-perfect scores capture the extent to which the paths deviate as described above. Based on these previous prompt templates and their performance metrics: \predvar{\{prompt\_history\}} Generate a new prompt template that might perform better. Focus on: 1. What worked well in high-performing templates 2. Avoiding patterns from low-performing templates 3. Recall the task is to identify birds without providing false information, while still trying to be specific (A high HP=1 is prioritized, then when it is close to 1 we want to improve HR) Return only the new template text, without any explanations.}
And each entry in \predvar{prompt\_history} is given by \modelpred{Template: \predvar{\{prompt\}} Hierarchical Recall (HR): \predvar{\{hr\}} Hierarchical Precision (HP): \predvar{\{hp\}}"} \\
Accuracy aware & \modelpred{You are an AI assistant helping to generate effective prompt templates for vision-language models to identify birds in images. The prompt templates should help the model provide accurate species identification that don't contain wrong information. Analyze the previous prompts and their performance metric (acc: how often the model's prediction is correct on average over the dataset) to generate an improved template which produces as high an accuracy as possible. Based on these previous prompt templates and their performance metrics: \predvar{\{prompt\_history\}} Generate a new prompt template that might perform better. Focus on: 1. What worked well in high-performing templates 2. Avoiding patterns from low-performing templates 3. Recall the task is to identify birds without providing false information, while still trying to be specific Return only the new template text, without any explanations.} And each entry in \predvar{prompt\_history} is given by \modelpred{Template: \predvar{\{prompt\}} Hierarchical Accuracy: \predvar{\{acc\}}}. \\
\bottomrule
\end{tabularx}
\caption{\textbf{System prompts for the bird classifier example.} These are the system prompts used to prompt ChatGPT to iterate on the classification prompt.}
\label{tab:sysprompts}
\end{table*}

We use ChatGPT to iterate on prompts for classifying birds without giving wrong information as described in \cref{sec:birdclass}. The model is given a ``system prompt'' that describes its task. These prompts are given in \cref{tab:sysprompts}\looseness=-1

\begin{table*}[t]
\small
\centering
\begin{tabularx}{\textwidth}{>{\raggedright\arraybackslash}m{2.5cm} >{\raggedright\arraybackslash}X}
\toprule
Model & Template \\
\midrule
\modelname{GPT-4} & \modelpred{\predvar{\{\}} Answer in the format A: \textless answer\textgreater.} \\
\modelname{LLaVA} & \modelpred{\predvar{\{\}} Answer in the format A: \textless answer\textgreater.} \\
\modelname{Fuyu} & \modelpred{Q: \predvar{\{\}} A:} \\
\modelname{ILXC2} & \modelpred{\textless ImageHere\textgreater \predvar{\{\}} Answer in the format A:  \textless answer\textgreater,}\\
\modelname{OmniLMM12B} & \modelpred{\predvar{\{\}} Answer in the format A: \textless answer\textgreater.} \\
\modelname{OmniLMM3B} & \modelpred{\predvar{\{\}} Answer in the format A: \textless answer\textgreater.} \\
\modelname{QwenVL} & \modelpred{Q: \predvar{\{\}} A:} \\
\bottomrule
\end{tabularx}
\caption{\textbf{Barebone prompt templates.} An overview of the prompts used with the various VLMs.}
\label{tab:sup_prompts2}
\end{table*}

\begin{table*}[t]
\small
\centering
\begin{tabularx}{\textwidth}{lX}
\toprule
Model & Template \\
\midrule
\modelname{GPT-4} & \modelpred{\predvar{\{\}} Do not give any extra text. Do not answer in a full sentence. Do not specify your certainty about the answer. Give your best guess if you are not sure. Be as specific as possible. Answer in the format A: \textless answer\textgreater.} \\
\modelname{LLaVA} & \modelpred{\predvar{\{\}} Do not give any extra text. Do not answer in a full sentence. Do not specify your certainty about the answer. Give your best guess if you are not sure. Be as specific as possible. Answer in the format A: \textless answer\textgreater.} \\
\modelname{Fuyu} & \modelpred{Q: \predvar{\{\}} A:} \\
\modelname{ILXC2} & \modelpred{\textless ImageHere \textgreater \predvar{\{\}} Do not give any extra text. Do not answer in a full sentence. Do not specify your certainty about the answer. Give your best guess if you are not sure. Be as specific as possible. Answer in the format A: \textless answer\textgreater.} \\
\modelname{OmniLMM12B} & \modelpred{\predvar{\{\}} Do not give any extra text. Do not answer in a full sentence. Do not specify your certainty about the answer. Give your best guess if you are not sure. Be as specific as possible. Answer in the format A: \textless answer\textgreater.} \\
\modelname{OmniLMM3B} & \modelpred{\predvar{\{\}} Do not give any extra text. Do not answer in a full sentence. Do not specify your certainty about the answer. Give your best guess if you are not sure. Be as specific as possible. Answer in the format A: \textless answer\textgreater.} \\
\modelname{QwenVL} & \modelpred{Q: \predvar{\{\}} A:} \\
\bottomrule
\end{tabularx}
\caption{\textbf{Specific prompt templates.} An overview of the prompts used with the various VLMs with the goal of providing a specific answer.}
\label{tab:sup_prompts3}
\end{table*}

\section{Example predictions and positions}
Some example VLM answers, mapped taxonomic predictions, and ground truth reference labels of the images are given in \cref{tab:location_examples_inat}. The prediction on the taxonomy is obtained using \cref{alg:tax_mapping}.

\begin{table*}[t]
\small
\centering
\label{tab:location_examples_inat}
\begin{tabular}{lm{4cm}m{4cm}m{4cm}}
\toprule
Model & Reference & VLM Answer & Predicted Node \\
\midrule
\modelname{LLaVA} & \concept{Strawberry Anemone} & \modelpred{Anemone} & \concept{Actiniid Sea Anemones} \\
\modelname{OmniLMM12B} & \concept{Velvety Tree Ant} & \modelpred{ant} & \concept{Carpenter Ants, Typical Sugar Ants and Allies} \\
\modelname{GPT-4} & \concept{Carex pilosa} & \modelpred{Equisetum (horsetail) species} & \concept{horsetails} \\
\modelname{QwenVL} & \concept{Nuttall's Snapdragon} & \modelpred{} & \concept{Animals} \\
\modelname{QwenVLChat} & \concept{sycamore maple} & \modelpred{maple} & \concept{maples} \\
\modelname{LLaVA} & \concept{Canadian wood nettle} & \modelpred{Plant} & \concept{dicots} \\
\modelname{OmniLMM3B} & \concept{Oysterplant} & \modelpred{Aloe vera} & \concept{aloe vera} \\
\modelname{Fuyu} & \concept{Cuckoo-pint} & \modelpred{This plant is a banana plant} & \concept{gingers, bananas, and allies} \\
\modelname{OmniLMM3B} & \concept{Familiar Bluet} & \modelpred{dragonfly} & \concept{Skimmers} \\
\modelname{OmniLMM3B} & \concept{Texan Crescent} & \modelpred{butterfly} & \concept{Brush-footed Butterflies} \\
\modelname{Fuyu} & \concept{Yellow-shouldered Slug Moth} & \modelpred{moth} & \concept{Butterflies and Moths} \\
\modelname{InternLmXC2} & \concept{Meadow Pipit} & \modelpred{sparrow} & \concept{New World Sparrows} \\
\modelname{OmniLMM12B} & \concept{Chihuahuan Nightsnake} & \modelpred{snake} & \concept{Colubrid Snakes} \\
\modelname{InternLmXC2} & \concept{fall phlox} & \modelpred{Asteraceae} & \concept{sunflowers, daisies, asters, and allies} \\
\modelname{Fuyu} & \concept{Crested Pigeon} & \modelpred{This is two pigeons} & \concept{Pigeons and Doves} \\
\modelname{OmniLMM12B} & \concept{Drummond's rain lily} & \modelpred{tulip} & \concept{monocots} \\
\modelname{QwenVL} & \concept{Strawberry Poison-dart Frog} & \modelpred{Strawberry poison frog} & \concept{Animals} \\
\modelname{LLaVA} & \concept{Red-masked Parakeet} & \modelpred{Parrot} & \concept{New World and African Parrots} \\
\modelname{Fuyu} & \concept{Hutton's Vireo} & \modelpred{bird} & \concept{Perching Birds} \\
\modelname{OmniLMM3B} & \concept{honewort} & \modelpred{ivy} & \concept{dicots} \\
\modelname{OmniLMM3B} & \concept{heath wood-rush} & \modelpred{grass} & \concept{grasses} \\
\modelname{QwenVL} & \concept{Cheetah} & \modelpred{cheetah} & \concept{Cheetah} \\
\modelname{LLaVA} & \concept{Gray Petaltail} & \modelpred{Dragonfly} & \concept{Skimmers} \\
\modelname{InternLmXC2} & \concept{Pima Rhatany} & \modelpred{This is a purple flower} & \concept{dicots} \\
\modelname{QwenVLChat} & \concept{Margined Leatherwing Beetle} & \modelpred{bug} & \concept{True Bugs, Hoppers, Aphids, and allies} \\
\modelname{OmniLMM12B} & \concept{Brazos rain-lily} & \modelpred{grass} & \concept{grasses} \\
\modelname{OmniLMM12B} & \concept{The Gem} & \modelpred{Lepidoptera} & \concept{Butterflies and Moths} \\
\modelname{GPT-4} & \concept{Blue Crane} & \modelpred{Crane} & \concept{Pelicans, Herons, Ibises, and Allies} \\
\modelname{QwenVLChat} & \concept{Twice-stabbed Stink Bug} & \modelpred{bugs} & \concept{True Bugs, Hoppers, Aphids, and allies} \\
\modelname{QwenVLChat} & \concept{Nacoleia rhoeoalis} & \modelpred{butterfly} & \concept{Brush-footed Butterflies} \\
\modelname{InternLmXC2} & \concept{White Bream} & \modelpred{This is a school of fish swimming in the ocean} & \concept{Perch-like Fishes} \\
\modelname{QwenVL} & \concept{Texan Crescent} & \modelpred{This is a butterfly} & \concept{Brush-footed Butterflies} \\
\modelname{OmniLMM12B} & \concept{Siberian Iris} & \modelpred{Iris} & \concept{Irises} \\
\modelname{QwenVL} & \concept{long beech fern} & \modelpred{Polystichum acrostichoides} & \concept{Christmas fern} \\
\modelname{QwenVL} & \concept{Black-mandibled Toucan} & \modelpred{Toucan} & \concept{Toucans} \\
\modelname{QwenVL} & \concept{Hazel Grouse} & \modelpred{Ruffed Grouse} & \concept{Ruffed Grouse} \\
\modelname{OmniLMM12B} & \concept{Hickory Tussock Moth} & \modelpred{caterpillar} & \concept{Underwing, Tiger, Tussock, and Allied Moths} \\
\modelname{QwenVL} & \concept{Goatsbeard} & \modelpred{Actaea pachypoda} & \concept{doll's eyes} \\
\modelname{Fuyu} & \concept{Definite Tussock Moth} & \modelpred{This caterpillar is a species of moth} & \concept{Butterflies and Moths} \\
\end{tabular}
\caption{\textbf{Examples of model predictions on the iNaturalist21 dataset.} `Model' refers to the model used to answer the question given an image. `VLM Answer' shows the output from the model, `Reference' is the true label, and `Predicted Node' is the node label we get from mapping onto the taxonomy using \cref{alg:tax_mapping}.}
\end{table*}

\section{Hierarchical Labels in Visual Recognition}
\label{sec:sem_sim_hiearchy}
Hierarchical labels have been studied for different aspects of visual representation learning, including contrastive representation learning~\cite{cole2022does}, weakly-supervised object localization~\cite{cole2022label}, open-set recognition~\cite{vaze2022openset,lang2024coarse}, and category discovery~\cite{zhao2024labeled,rastegar2024selex}.
A well-studied use case of hierarchical labels is to inform the learning of recognition models to reduce the `severity' of mistakes~\cite{hedging,bertinetto2020making,dhall2020hierarchical,kumar2017hierarchical,turkoglu2021crop,de2021digital}, e.g.,~by directly optimizing for the accuracy-specificity trade-off~\cite{hedging}, allowing a model to be accurate at the cost of specificity.
Similarly, incorporating knowledge graphs as expert-level human judgment has been shown to be beneficial \cite{marino2016know}.
Hierarchical recognition has also been explored in an open-world setting~\cite{verma2012learning,simeone2017hierarchical,lee2018hierarchical,bennette2021hierarchical,Dengxiong_2023_WACV} in which the goal is to relate novel, unseen categories to the training categories, e.g.,~by placing unseen categories on the taxonomy~\cite{verma2012learning}, or find the closest common taxonomic ancestor between the training data and the out-of-distribution sample~\cite{bennette2021hierarchical}.
Taxonomic information has also been used to speed up annotation \cite{gvanhorn2018lean}.
While not relying on explicit taxonomies, a taxonomic structure has also been extracted from text to enrich vision and language datasets to benefit contrastive learning~\cite{10204578}.
Computing semantic similarity between concepts or entities in taxonomies and knowledge graphs has been widely studied~\cite{resnik-1992-class, resnik-1995-infcontent, info_sim, comp_sem_kg, sem_sim_entity_dis}.
Most methods quantify similarity based on the information encoded in graph or taxonomy nodes, often using empirical probabilities to weight graph edges.
More recent approaches~\cite{comp_sem_kg, doi:10.1177/01655515211020580} incorporate corpus statistics alongside hierarchical structures.
In this work, we rely on unweighted node distances, as we focus on fine-grained classification domains where corpus frequencies are sparse and potentially unreliable.

\end{document}